\documentclass[11pt]{article}
\usepackage[a4paper]{geometry} 
\usepackage{clic2023} 
\usepackage{times} 
\usepackage{xurl} 
\usepackage[italian,english]{babel}
\usepackage{latexsym} 
\usepackage{xcolor}
\usepackage{comment}
\usepackage{graphicx} 
\usepackage[utf8]{inputenc}
\usepackage{breqn}
\usepackage{amsmath}
\usepackage{url}
\usepackage{microtype}
\usepackage{comment}
\usepackage{hyperref}
\usepackage{soul}
\usepackage{epstopdf}
\usepackage{booktabs}

\title{MedMine: Examining Pre-trained Language Models on Medication Mining}

\author{
           Haifa Alrdahi $^{1*}$,  Lifeng Han $^{1*}$, Hendrik Šuvalov $^{2*}$,  \and \textbf{Goran Nenadic}$^{ 1}$ \\
         $^1$ The University of Manchester, UK  \\ 
         $^2$ The University of Tartu, Estonia \\
         $*$ co-first authors
         \\ {\tt firstname.lastname @ \{manchester.ac.uk | ut.ee\}} \\
         }

\date{}

\begin{document}
\maketitle
\begin{abstract}
Automatic medication mining from clinical and biomedical text has become a popular topic due to its real impact on healthcare applications and the recent development of powerful language models (LMs).
However, fully-automatic extraction models still face obstacles to be overcome such that they can be deployed directly into clinical practice for better impacts. Such obstacles include their imbalanced performances on different entity types and clinical events.
In this work, we examine current state-of-the-art pre-trained language models (PLMs) on such tasks, via fine-tuning including the monolingual model Med7 and multilingual large language model (LLM) XLM-RoBERTa.
We compare their advantages and drawbacks using historical medication mining shared task data sets from n2c2-2018 challenges. 
We report the findings we get from these fine-tuning experiments such that they can facilitate future research on addressing them, for instance, how to combine their outputs, merge such models, or improve their overall accuracy by ensemble learning and data augmentation. MedMine is part of the M3 Initiative \url{https://github.com/HECTA-UoM/M3}
\end{abstract}

\section{Introduction}

Medication mining plays a vital role in clinical natural language processing (ClinicalNLP) applications and digital healthcare settings. 
For instance, the extraction of medications used in patients' historical electronic health records and the corresponding targeted treatments can be very important features for the cohort selection of certain diseases and treatments. 
Medications and corresponding adverse drug effects extraction can be  studied for future optimised and personalised treatments.
Meditation extraction itself can also be beneficial to epidemiological studies and term management \cite{Term_mana_biome2006}, e.g. to rheumatologists, and the extracted medical terminologies and concepts can be useful for knowledge transformation \cite{han-etal-2023-investigating,han-etal-2022-examining}. 

Medication extraction has been an application field of NLP models for decades across statistical and neural NLP methods \cite{10.1136/jamia.2010.003657_MedIE,10.1093/jamia/ocy114_MedNorm}.
With the breakthrough of advanced learning structures based on Transformer and pre-trained language models (PLMs) BERT \cite{devlin-etal-2019-bert}, researchers have reported new results on clinical terminology mining. 
In this project, we aim at re-examining current state-of-the-art medication mining and large language models (LLMs) on medication extraction tasks by carrying out fine-tunings. 
We investigate the strengths and weaknesses of these models, explore the possibility to unify or merge their outputs, and have the goal in mind to develop an augmented medication mining toolkit and platform for open research, which we name as \textbf{MedMine}.

To the best of our knowledge, MedMine is the first attempt (or one of the first) to integrate existing state-of-the-art information extraction and LLMs in the healthcare and clinical domain, especially on medication extraction tasks.

\section{Related Work}

Some relevant work to ours includes the following categories: 1) \textit{Medication mining in social medial} data to monitor medication abuses.
For instance,  \newcite{sarker2016social_med} investigated the possibility of drug overuse detection using Twitter user posts from the USA on medications including Adderall, oxycodone, and quetiapine, in comparison to controlled medication metformin. They annotated 6,400 tweets manually and trained the classifier using LibSVM and Weka toolkit, which achieved around 46\% of the F1 score on prediction.

2) \textit{Medication mining to identify risks} and improve treatments. For instance, \newcite{harkanen2019identifying_risk} used around 70K medication incident reports from England and Wales and conducted relevant events mining around the risk areas. The outcomes of identified areas included allergic reactions, intravenous administration of antibacterial drugs, and Fentanyl patches, etc. that need more attention.
\newcite{bereznicki2008dataMining_asthma} carried out case studies by looking into asthma patients' medication records using data mining software applications from Australia. Insightful observations were achieved from statistical analysis and intervention letters were sent to patients to visit their GP.

3) \textit{Prediction of next prescribed medications}. \newcite{wright2015use_sequential_predict} took diabetes patients' medication records and carried out the next prescribed medication prediction task using sequential pattern mining and cSPADE toolkit \cite{buchta2023package}. Out of 161,497 patients' data, their experiment achieved around 90\% evaluation scores on the drug class level and 64\% at the generic drug level. This work also indicated the usefulness of temporal information in relationship modelling of medications. 

4) \textit{Integrating NLP models for healthcare}. Instead of medication mining, MedCAT is a platform developed by \newcite{Kraljevic2021-MedCAT} focusing on diagnoses extraction tasks from clinical text. The integrated models inside MedCAT \footnote{\url{https://github.com/CogStack/MedCAT}} include statistical ones such as CRF, and neural models such as Transformers, as well as their combinations. MedCAT also has an entity linking function to normalise the extracted entities into existing clinical terminologies databases such as SNOMED CT and UMLS.

Some very relevant shared tasks and workshops (WS) include the n2c2 challenge series \footnote{\url{https://n2c2.dbmi.hms.harvard.edu}} which data we will use in this paper for the experimental evaluations, the Bio-medicine and its Applications NLP WS (BioNLP) \cite{ws-2004-international_bionlp,bionlp-2023-biomedical}, Louhi WS on Text and Data Mining of Health Documents \cite{louhi-2022-international,ws-2010-naacl-hlt-2010-louhi}. 
In the UK, we also organised the HealTAC conference series on Healthcare Text Analytics \footnote{\url{http://healtex.org} 6th Edition this year hosted in the University of Manchester}.




\section{Methodologies and Experiments}
In this section, we introduce \textbf{MedMine}, the first version of our Medication Mining Integration project.

\begin{figure}[!t]
\begin{center}
\centering
\includegraphics*[width=0.49\textwidth]{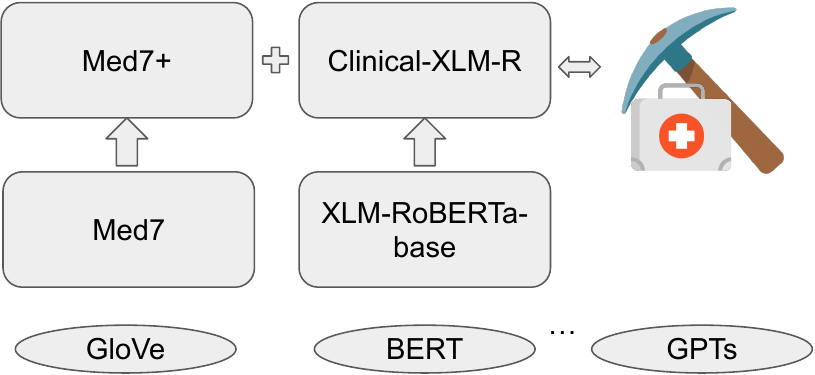}
\caption{MedMine Illustration: Models already included Glove and RoBERTa embeddings, clinical domain fine-tuning for medication mining.}
\label{fig:MedMine-diagram}
\end{center}
\end{figure}

In MedMine.V0, we try to explore the current large language models (LLMs), especially BERT-based models \cite{devlin-etal-2019-bert} in comparison to pre-BERT embeddings. The two models we chose are 1) Med7 \cite{KORMILITZIN2021102086_med7} which is a named entity recognition (NER) model fine-tuned in clinical domain medical records using SpaCy's library of GloVe embedding \cite{pennington-etal-2014-glove}, and 2) XLM-RoBERTa-base which is a multilingual pre-trained RoBERTa model \cite{conneau-etal-2020-unsupervised_roberta} using 100 languages from CommonCrawl corpora. The goal is to achieve fine-tuning on clinical domain data for medication mining tasks as demonstrated in Figure \ref{fig:MedMine-diagram}.

\subsection{Models and Data}

The two off-the-shelf models we deployed in our experiments are from HugingFace XLM-RoBERTa-base \footnote{\url{https://huggingface.co/xlm-roberta-base}} and Github Med7 \footnote{\url{https://github.com/kormilitzin/med7}}.
For fine-tuning data, we used the n2c2-2018 track-2 shared task corpus which originally includes 303 letters for training and 202 letters for testing. The corpus was annotated manually by four physician assistant students and three nurses \cite{henry20202018_n2c2_task2}. The corresponding sentence counts are 46,033 and 30,614 from the original training and testing set \cite{wu2022crossdomain}.

This shared task is on Adverse Drug Events and Medication Extraction in Electric Health Records (EHRs).
To prepare a new data setting with a validation set, we split the original 505 annotated letters into 70/15/15\% for training, development, and testing which corresponds to the following set (353, 76, 76).


\subsection{Fine-tuning Parameters}

For Med7, the fine-tuning was using 30 iterations (epochs), and for XML-R-Base, the fine-tuning parameters are displayed below.
\begin{itemize}
    \item batch size = 16
    \item evaluation strategy = ``epoch",
    \item learning rate=1e-4,
    \item num train epochs=8,
    \item weight decay=1e-5
    \item metric = load metric(``seqeval")
    \item tokenizer = true
\end{itemize}

\subsection{Med7+ Outputs}

Med7 has only 7 labels in the outputs including Dosage, Drug, Duration, Form, Frequency, Route, and Strength, but our fine-tuned Med7+ has 9 labels fine-tuned from n2c2-2018 shared task with ADE and Reason labels.

Med7 Deployment using the 7 labels tested on 76 letters is evaluated in  Table \ref{tab:med7-deploy-76} with 
              precision,   recall,  f1-score, micro avg, macro avg, and weighted avg. The ``support'' column is the number of real labels in the reference. The evaluation score shows that Med7 has really low performances on the `dosage' label achieving 0.11 for Precision, 0.24 for Recall, and 0.15 for F1, even though the number of true labels on dosage (1039) is not much smaller than most of the other labels.
              
\begin{table}[!h]
\begin{center}
\centering
\begin{tabular}{crccc}
\toprule
\multicolumn{1}{c}{Catergory} 
     & \multicolumn{1}{c}{Pre. }     
                & Rec.   & F1   & Support \\
\midrule

        form   &    0.90  &    0.90  &    0.90  &    1696\\
    strength   &    0.70  &    0.80  &    0.75  &    1639\\
      dosage     &  0.11   &   0.24     & 0.15   &   1039\\
        drug     &  0.90  &    0.77  &    0.83  &    3954\\
       route    &   0.96    &  0.94    &  0.94   &   1341\\
   frequency     &  0.74  &    0.79   &   0.76  &    1564\\
    duration   &    0.73    &  0.75    &  0.74     &  139\\ \hline 
   micro avg  &     0.71   &   0.77   &   0.74 &    11372\\
   macro avg  &     0.72  &    0.74   &   0.72  &   11372\\
weighted avg   &    0.78    &  0.77   &   0.77  &   11372\\
\bottomrule
\end{tabular}
\caption{Outputs from Med7 Deployment on 76 Testing Letters - Evaluation: Type}
\label{tab:med7-deploy-76}
\end{center}
\end{table}

\begin{table}[!h]
\begin{center}
\centering
\begin{tabular}{crccc}
\toprule
\multicolumn{1}{c}{Catergory} & \multicolumn{1}{c}{Pre. }                   & Rec.   & F1   & Support \\
\midrule
           O  &   0.0000 &   0.0000  &  0.0000   &    874\\
      reason  &   0.7276 &   0.4552 &   0.5601   &    927\\
         ade  &   0.5579  &  0.2190 &   0.3145  &     242\\
        form  &   0.9229 &   0.9393  &  0.9310  &    1696\\
    strength  &   0.9749  &  0.9494 &   0.9620  &    1639\\
      dosage  &   0.9124  &  0.8816 &   0.8967 &     1039\\
        drug  &   0.9345  &  0.9135 &   0.9239   &   3954\\
       route  &   0.9580 &   0.9366 &   0.9472  &    1341\\
   frequency   &  0.8502  &  0.9399  &  0.8928   &   1564\\
    duration    & 0.8015   & 0.7554   & 0.7778    &   139\\ \hline 
    accuracy     &&&                    0.8187     &13415\\
   macro avg   &  0.7640   & 0.6990    &0.7206     &13415\\
weighted avg    & 0.8454   & 0.8187    &0.8282     &13415\\ \hline
\multicolumn{5}{c}{Removing label `O': only using 9 labels}  \\\hline 
   micro avg   &  0.9124 &   0.8758 &   0.8937 &    12541\\
   macro avg    & 0.8489  &  0.7767  &  0.8007  &   12541\\
weighted avg     &0.9044   & 0.8758   & 0.8859   &  12541\\
\hline 
\multicolumn{5}{c}{Removing labels `O', `Reason', and `ADE':}  \\
\multicolumn{5}{c}{check the improvement on 7 labels}  \\\hline 
   micro avg  &   0.9248 &   0.9240 &   0.9244 &    11372\\
   macro avg   &  0.9078  &  0.9022  &  0.9045  &   11372\\
weighted avg    & 0.9262   & 0.9240   & 0.9247   &  11372\\

\bottomrule
\end{tabular}
\caption{Outputs from Med7+ fine-tuning using 9labels on 76 Letters - Evaluation: Type}
\label{tab:med7-fine-tune-9plusO-76}
\end{center}
\end{table}

The fine-tuned Med7+ is reported in Table \ref{tab:med7-fine-tune-9plusO-76} using the same test set where it gives 9 labels. 
Furthermore, for the existing 7 labels, the Med7+ produced much higher scores than the Med7 baseline model across most of the labels, boosting the individual F1 scores up to around 0.90+ except for the `duration' label, which is improved from 0.74 to 0.78 (the first section of Table \ref{tab:med7-fine-tune-9plusO-76}). The bottom two sections of Table \ref{tab:med7-fine-tune-9plusO-76} analyse the evaluation scores when we remove the `O' label and also the `Reason' and `ADE' labels. The bottom score by removing all three labels is to compare our fine-tuned Med7+ directly to the original Med7 baseline on the 7 labels, which says that we have improved the micro, macro, and weighted avg scores all from 0.70s to 0.90s.

Both Table \ref{tab:med7-deploy-76} and Table \ref{tab:med7-fine-tune-9plusO-76} reported the evaluation using ``Lenient Matching'' which corresponds to the ``Type'' category out of the four 
different Evaluation Strategies introduced from SemEval2013 \cite{semeval-2013-joint-lexical} \footnote{\url{https://paperswithcode.com/dataset/semeval-2013}}.
\begin{itemize}
    \item {Strict}: ``exact boundary surface string match and entity type'';
    \item Exact: ``exact boundary match over the surface string, regardless of the type'';
    \item Partial: ``partial boundary match over the surface string, regardless of the type'';
    \item \textbf{Type}: ``some overlap between the system tagged entity and the gold annotation is required''.
\end{itemize}


\subsection{Clinical-XLM-R Outputs}
We name the fine-tuned XLM-RoBERTa-base as Clinical-XLM-R which produces all the 9 labels output as shown in Table \ref{tab:clinical-xlm-r} using the same evaluation strategy we used for Med7+ ``Lenient Matching''. 

\begin{table}[!h]
\begin{center}
\centering
\begin{tabular}{crccc}
\toprule
\multicolumn{1}{c}{Acc} 
     & \multicolumn{1}{c}{Pre }     
                & Rec   & F1   &  \\
\midrule
96.76\% & 87.98\%    & 90.14\% & 89.05\% &     \\
\hline
\multicolumn{1}{c}{Catergory} 
     & \multicolumn{1}{c}{Pre. }     
                & Rec.   & F1   & Num. \\
\midrule
ADE & 51.96\%    & 55.09\% & 53.48\% & 432     \\
Dosage & 85.94\%    & 91.40\% & 88.59\% & 803    \\ 
Drug & 93.05\%   & 95.63\%  & 94.32\% & 8193 \\ 
Duration & 58.40\%    & 67.34\% & 62.55\% & 98   \\ 
Form & 91.69\%    & 90.46\% & 91.07\% & 1647   \\  
Frequency & 88.13\%    & 90.09\% & 89.10\% & 1838    \\  
Reason & 59.57\%    & 62.01\% & 60.77\% & 1369    \\  
Route & 90.90\%    & 91.28\% & 91.09\% & 1205  \\  
Strength & 94.83\% & 96.00\% & 95.41\% & 1377   \\  
\bottomrule
\end{tabular}
\caption{Outputs from Clinical-XLM-RoBERTa}
\label{tab:clinical-xlm-r}
\end{center}
\end{table}

In Table \ref{tab:clinical-xlm-r}, the first row is the overall Accuracy, Precision, Recall, and F1 score. The column 'Num.' is the number of entities the model predicted. 
Some interesting findings from Clinical-XLM-R are listed below: 
\begin{itemize}
    \item 1) ADE, Duration, and Reason are the three most challenging categories to identify, especially since their precision scores are under 60\% while others are around 90\%.
    \item 2) Most categories have relatively higher Recall than Precision scores except for `Form' labels, which means the system has more false positive labels than false negative labels. One of the future directions to optimise the model is to assign more attention to restrict false positive labels. However, this also depends on the practical application of medication and related event extraction. For instance, if a High Recall score is preferred to identify all useful information, the model is doing a good job at his stage.
    \item 3) The tokenizer we used for this model is from the XLM-RoBERTa-base package. We handled documents that were larger than 512 by splitting them up and dividing them into 512-sized chunks. The downside of this is that it might lose some context information from the beginning of the document.
\end{itemize}

\section{Discussion}
Looking into both Med7+ and Clinical-XLM-R regarding this specific task on Drugs and related events extraction using n2c2-2018 challenge data, there are some research directions worthy to be explored. 
\begin{itemize}
    \item 1) How to address the label imbalance issue? e.g. Duration and ADE have apparent much lower rates of labels (139, 242) out of 13,415, while other labels have around 1K labels, in addition to `Drug' having around 4K labels. Data augmentation via oversampling or synthetic data generation can be a possible solution toward a more balanced data setting, which hopefully can improve model accuracy on those low-frequency label predictions.
    \item 2) Med7 has kind of balanced Precision and Recall scores. While Med7+ improved Med7 scores in each category except for the label `route' which stayed almost the same scores (0.96->0.96, 0.94->0.94, 0.94-0.95), the Precision scores of Med7+ are in general higher than the Recall scores except for `form' and `frequency'.
    This is contradictory to Clinical-XLM-R, which has in general higher Recall scores instead. It is our next step to look into the fine-tuning details of these two models, and hopefully, we can enhance both models from each other's advantages.
    \item 3) Clinical-XLM-R boosted the overall Accuracy score to 96.76, in comparison to 81.87 from Med7+. This might indicate that the BERT-based word embeddings learn can learn better performance than GloVe embeddings on this task; however, it is also possible that the multilingual training data XLM-RoBERTa used is helpful for pre-training, even though Med7 has been pre-trained on their available clinical domain data-set. We need to look into this in detail for model explanation.
\end{itemize}

\section{Conclusion and Future Work}
As part of the M3 Initiative \cite{M3}, we reported the MedMine project progress using clinical domain fine-tuning of two PLMs i.e. Med7 and XLM-R. Med7 itself is already a fine-tuned model using clinical domain data that reports 7 labels, while XLM-RoBERTa is a multilingual pre-trained general domain large language model (LLM). 
The outcomes of Med7+ and Clinical-XLM-R both demonstrated much higher performances than the Med7 baseline model across most of the 9 original labels using n2c2-2018 shared task data. 
Comparisons restricted on the 7 labels reported by Med7, Med7+ produced F1 scores (0.9244, 0.9045, 0.9247) vs (0.74, 0.72, 0.77) from Med7 model, for micro avg, macro avg, and weighted avg using our experimental setup. 

In future work, we plan to look into example outputs with expert humans regarding different models and investigate the strategies on merging their outputs. 
We also have an interest in expanding integrated models into GPTs as in Figure \ref{fig:MedMine-diagram}, and explore different prompt-based learning (PBL) mechanisms for medication extraction using both manual and soft templates, such as mixed template used by \newcite{cui-etal-2023-medtem2} for temporal modelling of medications.

\section*{Limitations}
In this work, we used n2c2-2018 shared task data, which limits the size of the training and testing set. It is our plan for the next steps to extend the training set and label categories e.g. `Temporal' by combining more historical n2c2/i2b2 challenge data such as the ones from 2009 and 2012. This will involve merging annotated data sets  taking into account their differences in formats. 

\section*{Ethical Concerns}
The data set we used from the n2c2 challenge is anonymised and we have gone through good clinical practice training to access the data, without sharing them with any third parties.

\section*{Acknowledgments}
We thank Arooj Hussain, Mingyang Li, Yuping Wu, and Warren Del-Pinto for their input on the ongoing M3 project and valuable discussion.
This project is partially funded by the University of Manchester Open Research Office via OR Fund Project ``An Open-Research Framework on Label Augmentations for Low-Resource Clinical Natural Language Processing  using Graph-Based Semi-Supervised Learning'' and the UKRI/EPSRC grant EP/V047949/1 ``Integrating hospital outpatient letters into the healthcare data space''.

\section*{Author Contributions}
HA: Med7 deployment and fine-tuning; HS: XLM-R fine-tuning; LH: supervised the project and drafted the paper; 
GN: co-supervisor of MedMine and discussion.
\bibliographystyle{acl}
\bibliography{bibliography.bib}

\section*{Appendix}

Because Med7 itself is fine-tuned for n2c2-2018 shared task, we can deploy it directly to test the performances. Here we also list its output scores when we evaluate it using the official n2c2-2018 training data, i.e. the 303 labelled letters. Pay attention that ADE and Reason labels are missing from the model via its direct evaluation.

\begin{table}[!h]
\begin{center}
\centering
\begin{tabular}{crccc}
\toprule
\multicolumn{1}{c}{Catergory} 
     & \multicolumn{1}{c}{Pre. }     
                & Rec.   & F1   & Num. \\
\midrule
Dosage & 0.94    & 0.93 & 0.93 & 4221    \\ 
Drug & 0.93   & 0.89  & 0.91 & 16225 \\ 
Duration & 0.82    & 0.85 & 0.83 & 592   \\ 
Form & 0.94  & 0.92 & 0.93 & 6651   \\  
Frequency & 0.88    & 0.82 & 0.85 & 6281    \\  
Route & 0.96  & 0.96 & 0.96 & 5476  \\  
Strength & 0.93 & 0.94 & 0.94 & 6691   \\  \hline 
Acc. &&& 0.85& 49003\\
Macro avg & 0.80&0.79 & 0.79& 49003\\
Weighted avg &0.87 &0.85 & 0.86& 49003\\
\bottomrule
\end{tabular}
\caption{Outputs from Med7 Deployment on 303 Letters - Evaluation: Type}
\label{tab:med7-deploy-type}
\end{center}
\end{table}

\begin{table}[!t]
\begin{center}
\centering
\begin{tabular}{crccc}
\toprule
\multicolumn{1}{c}{Catergory} 
     & \multicolumn{1}{c}{Pre. }     
                & Rec.   & F1   & Num. \\
\midrule
Dosage & 0.94    & 0.91 & 0.92 & 4221    \\ 
Drug & 0.93   & 0.88  & 0.90 & 16225 \\ 
Duration & 0.81    & 0.82 & 0.82 & 592   \\ 
Form & 0.94  & 0.91 & 0.92 & 6651   \\  
Frequency & 0.88    & 0.77 & 0.82 & 6281    \\  
Route & 0.96  & 0.96 & 0.96 & 5476  \\  
Strength & 0.93 & 0.94 & 0.94 & 6691   \\  \hline 
Acc. &&& 0.83& 49003\\
Macro avg & 0.80&0.77 & 0.78& 49003\\
Weighted avg &0.87 &0.83 & 0.85& 49003\\
\bottomrule
\end{tabular}
\caption{Outputs from Med7 Deployment on 303 Letters - Evaluation: Strict}
\label{tab:med7-deploy-strict}
\end{center}
\end{table}

The `Frequency' catergory score has a big drop on Recall with absolute value 0.05 via `Strict' in comparison to `Type', otherwise, the recall scores mostly have 0.01-0.02 drops and Precision scores remain the same for other labels.

\end{document}